\documentclass[sigconf]{acmart}

\renewcommand\footnotetextcopyrightpermission[1]{} % removes footnote with conference information in first column
\usepackage{graphicx}
\usepackage{booktabs}
\usepackage{enumitem}
\usepackage{enumerate}
\usepackage{amssymb}
\usepackage{amsmath} % can use the split
\usepackage{textcomp}
\def\BibTeX{{\rm B\kern-.05em{\sc i\kern-.025em b}\kern-.08emT\kern-.1667em\lower.7ex\hbox{E}\kern-.125emX}}

\begin{document}

%
% The "title" command has an optional parameter, allowing the author to define a "short title" to be used in page headers.
\title[FiBiNET]{FiBiNET: Combining Feature Importance and Bilinear feature Interaction for Click-Through Rate Prediction}
% \title[Short title]{FiBiNET}

\author{Tongwen Huang}
% \authornote{Dr.~Trovato insisted his name be first.}
\orcid{1234-5678-9012}
\affiliation{%
  \institution{Sina Weibo Inc.}
}
\email{tongwen@staff.weibo.com}

\author{Zhiqi Zhang}
% \authornote{Equal contribution}
\affiliation{%
  \institution{Sina Weibo Inc.}
}
\email{zhiqizhang@staff.weibo.com}

\author{Junlin Zhang}
% \authornote{Equal contribution}
\affiliation{%
  \institution{Sina Weibo Inc.}
}
\email{junlin6@staff.weibo.com}

%
% By default, the full list of authors will be used in the page headers. Often, this list is too long, and will overlap
% other information printed in the page headers. This command allows the author to define a more concise list
% of authors' names for this purpose.
% The abstract is a short summary of the work to be presented in the article.
\begin{abstract}
Advertising and feed ranking are essential to many Internet companies such as Facebook and Sina Weibo. Among many real-world advertising and feed ranking systems, click through rate (CTR) prediction plays a central role. There are many proposed models in this field such as logistic regression, tree based models, factorization machine based models and deep learning based CTR models. However, many current works calculate the feature interactions in a simple way such as Hadamard product and inner product and they care less about the importance of features. In this paper, a new model named FiBiNET as an abbreviation for Feature Importance and Bilinear feature Interaction NETwork is proposed to dynamically learn the feature importance and fine-grained feature interactions. On the one hand, the FiBiNET can dynamically learn the importance of features via the Squeeze-Excitation network (SENET) mechanism; on the other hand, it is able to effectively learn the feature interactions via bilinear function. We conduct extensive experiments on two real-world datasets and show that our shallow model outperforms other shallow models such as factorization machine(FM) and field-aware factorization machine(FFM). In order to improve performance further, we combine a classical deep neural network(DNN) component with the
shallow model to be a deep model. The deep FiBiNET consistently outperforms the other state-of-the-art deep models such as DeepFM and extreme deep factorization machine(XdeepFM).
\end{abstract}


% \ccsdesc{\ccsdesc[300]{~Computational advertising}
% ~Robotics}


%
% A "teaser" image appears between the author and affiliation information and the body 
% of the document, and typically spans the page. 
%   \Description{Enjoying the baseball game from the third-base seats. Ichiro Suzuki preparing to bat.}

%
% This command processes the author and affiliation and title information and builds
% the first part of the formatted document.
\maketitle

\section{Introduction}

Advertising and feed ranking are essential to many Internet companies such as Facebook and Sina Weibo. The main technique behind these tasks is click-through rate prediction which is known as CTR. Many models
have been proposed in this field such as logistic regression (LR)\cite{mcmahan2013ad}, polynomial-2 (Poly2)\cite{juan2016field}, tree based models\cite{he2014practical}, tensor-based models\cite{koren2009matrix}, Bayesian models\cite{graepel2010web}, and factorization machines  based models\cite{rendle2010factorization,rendle2012factorization,juan2016field,juan2017field}.

With the great success of deep learning in many research fields such as
computer vision\cite{krizhevsky2012imagenet, he2016deep} and natural language processing\cite{mikolov2010recurrent, cho2014learning}, many
deep learning based CTR models have been proposed in recent
years\cite{zhang2016deep,cheng2016wide,xiao2017attentional,guo2017deepfm,lian2018xdeepfm,wang2017deep,zhou2018deep,He2017NFM}. As a result, the deep learning for CTR prediction has also been a research trend in this field. Some neural network based models have been proposed and achieved success such as Factorization-Machine Supported Neural Networks (FNN)\cite{zhang2016deep},
Wide\&Deep model(WDL)\cite{cheng2016wide}, Attentional Factorization
Machines(AFM)\cite{xiao2017attentional},  DeepFM\cite{guo2017deepfm}, XDeepFM\cite{lian2018xdeepfm} etc.

In this paper, a new model named FiBiNET as an abbreviation for Feature Importance and Bilinear feature Interaction NETwork is proposed to dynamically learn the feature importance and fine-grained feature interactions. As far as we know, different features have various importances for the target task. For example, the feature occupation is more important than the feature hobby
when we predict a person's income. Taking this into consideration, we introduce a Squeeze-and-Excitation network (SENET)\cite{hu2017squeeze} to learn the weights of features dynamically. Besides, feature interaction is a key challenge in CTR prediction field and many related works calculate the feature interactions in a simple way such as Hadamard product and inner product. We propose a new fine-grained way in this paper to calculate the feature interactions with the bilinear function.
Our main contributions are listed as follows:

\begin{itemize}
\item
  Inspired by the success of SENET in the computer vision field, we use the SENET mechanism to learn the weights of features dynamically.
\item
  We introduce three types of Bilinear-Interaction layer to learn
  feature interactions in a fine-grained way. This is also in contrast
  to the previous work\cite{juan2016field,juan2017field, xiao2017attentional, rendle2010factorization,rendle2012factorization,He2017NFM}, which calculates the feature interactions with Hadamard product or inner product.
\item
  Combining SENET mechanism with bilinear feature interaction, our shallow model achieves state-of-the-art among the shallow models such as FFM on Criteo and Avazu datasets.
\item
  For further performance gains, we combine a classical deep neural network(DNN) component with the shallow model to be a deep model. The deep FiBiNET consistently outperforms the other state-of-the-art deep models
  on Criteo and Avazu datasets.
\end{itemize}

The rest of this paper is organized as follows. In Section \ref{sec:s2}, we review
related works which are relevant with our proposed model, followed by
introducing our proposed model in Section \ref{sec:s3}. We will present
experimental explorations on Criteo and Avazu datasets in Section \ref{sec:s4}. Finally, we
discuss empirical results and conclude this work in Section \ref{sec:s5}. 

\section{Related Work}
\label{sec:s2}
\subsection{Factorization Machine and Its relevant variants}

Factorization machine(FM)\cite{rendle2010factorization,rendle2012factorization} and field-aware factorization machine (FFM)\cite{juan2016field,juan2017field} are two of the most successful CTR models. FM models all feature interactions between variables using factorized
parameters. It has a low time complexity and memory storage, and it
works well on large sparse data. FFM introduced field aware latent
vectors and won two competitions hosted by Criteo and Avazu\cite{juan2017field}.
However, FFM was restricted by the need of large memory and cannot
easily be used in Internet companies.
\subsection{Deep Learning based CTR Models}
Deep learning has achieved great success in many research fields
such as computer vision\cite{krizhevsky2012imagenet, he2016deep} and natural language processing\cite{mikolov2010recurrent, cho2014learning}. As a result, many deep learning based CTR models have also been proposed in recent years\cite{zhang2016deep,cheng2016wide,xiao2017attentional,guo2017deepfm,lian2018xdeepfm,wang2017deep,zhou2018deep,He2017NFM}. How to
effectively model the feature interactions is the key factor for most of
these neural network based models.

Factorization-Machine Supported Neural Networks (FNN)\cite{zhang2016deep} is a
forward neural network using FM to pre-train the embedding layer.
However, FNN can capture only high-order feature interactions. Wide \&
Deep model(WDL)\cite{cheng2016wide} was initially introduced for application
recommendation in google play. WDL jointly trains wide linear models and
deep neural networks to combine the benefits of memorization and
generalization for recommender systems. However, expertise feature
engineering is still needed on the input to the wide part of WDL, which
means that the cross-product transformation also requires to be manually
designed. To alleviate manual efforts in feature engineering,
DeepFM\cite{guo2017deepfm} replaces the wide part of WDL with FM and shares the
feature embedding between the FM and deep component. DeepFM is regarded
as one state-of-the-art model in CTR estimation field.

Deep \& Cross Network (DCN)\cite{wang2017deep} efficiently captures feature
interactions of bounded degrees in an explicit fashion. Similarly,
eXtreme Deep Factorization Machine (xDeepFM)\cite{lian2018xdeepfm} also models the
low-order and high-order feature interactions in an explicit way by
proposing a novel Compressed Interaction Network (CIN) part.

As \cite{xiao2017attentional} mentioned, FM can be hindered by its modeling of all feature
interactions with the same weight, as not all feature interactions are
equally useful and predictive. And they propose the Attentional
Factorization Machines(AFM)\cite{xiao2017attentional} model, which uses an attention
network to learn the weights of feature interactions. Deep Interest
Network (DIN)\cite{zhou2018deep} represents users' diverse interests with an
interest distribution and designs an attention-like network structure to
locally activate the related interests according to the candidate ad.

\subsection{SENET Module}
Hu \cite{hu2017squeeze} proposed the ``Squeeze-and-Excitation Network'' (SENET) to improve the representational power of a network by explicitly modeling
the interdependencies between the channels of convolutional features in
various image classification tasks. The SENET is proved to be successful
in image classification tasks and won first place in the ILSVRC 2017
classification task.

There are also other applications about SENET except for the image
classification\cite{yan2018weakly,kitada2018skin,roy2018recalibrating}. \cite{roy2018recalibrating} introduces three variants of the SE modules for semantic segmentation task. Classifying common thoracic diseases as
well as localizing suspicious lesion regions on chest X-rays\cite{yan2018weakly} is another application field. \cite{linsley2018global}
extends SENET module with a global-and-local attention (GALA) module
to get state-of-the-art accuracy on ILSVRC.

\section{Our Proposed Model}
\label{sec:s3}
\begin{figure*}
\centering
\includegraphics[width=14cm, height=8cm]{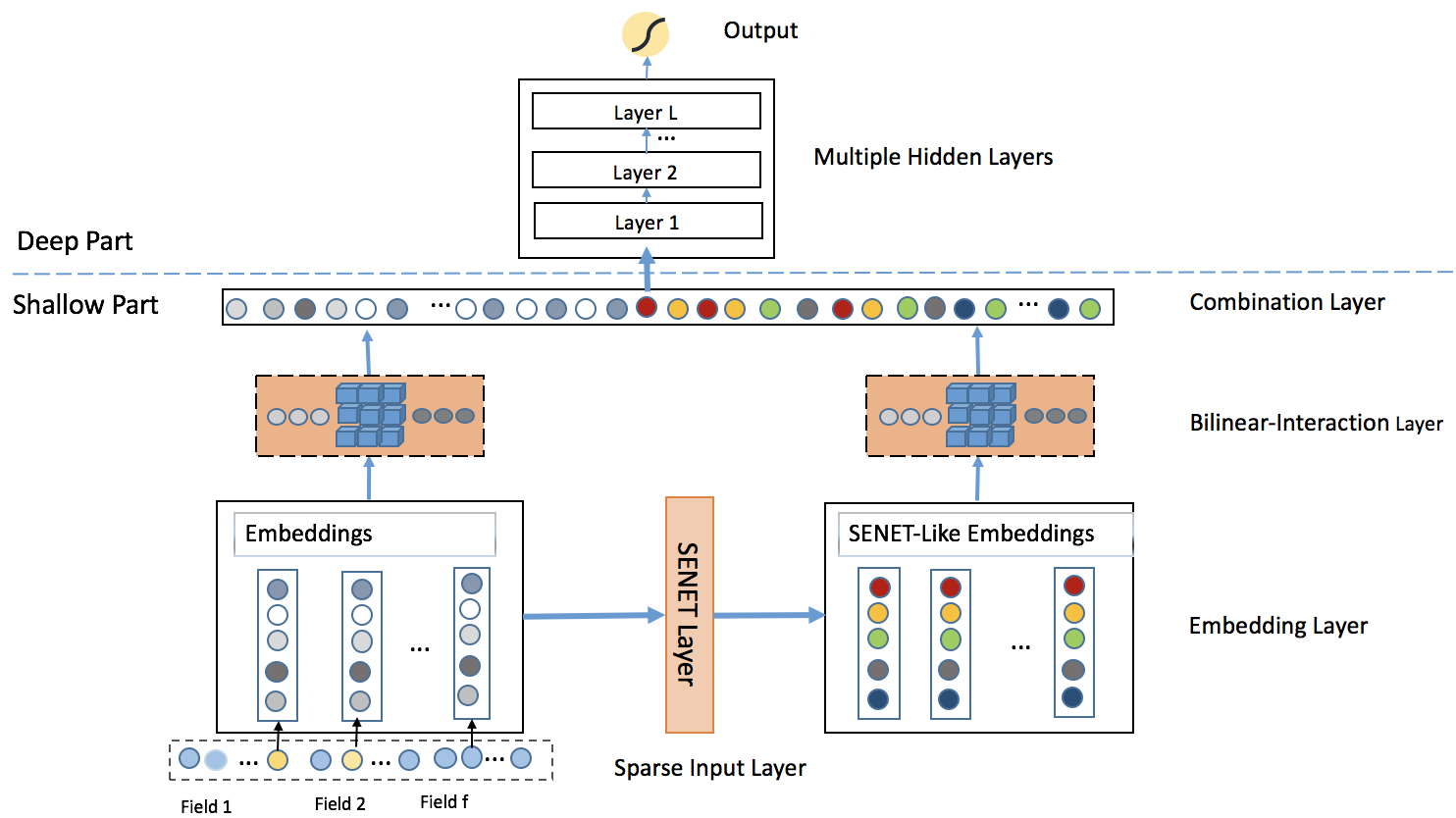}
\caption{The architecture of our proposed FiBiNET}
\label{fig:f1}
\end{figure*}
We aim to dynamically learn the importance of features and feature interactions in a
fine-grained way. To this end, we propose the Feature Importance and
Bilinear feature Interaction NETwork(FiBiNET) for CTR prediction tasks.

In this section, we will describe the architecture of our proposed model as depicted in Figure \ref{fig:f1}. For clarity purpose, we omit the logistic regression part which
can be trivially incorporated. Our proposed
model consists of the following parts: sparse input layer, embedding
layer, SENET layer, Bilinear-Interaction layer, combination layer,
multiple hidden layers and output layer. The sparse input layer and
embedding layer are the same with DeepFM\cite{guo2017deepfm}, which adopts a sparse
representation for input features and embeds the raw feature input into
a dense vector. The SENET layer can convert an embedding layer into the
SENET-Like embedding features, which helps to boost feature
discriminability. The following Bilinear-Interaction
layer models second order feature interactions on the original embedding and
the SENET-Like embedding respectively. Subsequently, these cross features are concatenated by a combination layer which merges the
outputs of Bilinear-Interaction layer. At last, we feed the cross features into a deep neural network and the network outputs the
prediction score.

\subsection{Sparse Input and Embedding
layer}
The sparse input layer and embedding layer are widely used in
deep learning based CTR models such as DeepFM\cite{guo2017deepfm} and AFM\cite{xiao2017attentional}.
The sparse input layer adopts a sparse representation for raw input
features. The embedding layer is able to embed the sparse feature into a
low dimensional, dense real-value vector. The output of embedding layer
is a wide concatenated field embedding\footnote{The field embedding is
  also known as the feature embedding. If the field is multivalent, the
  sum of feature embedding is used as the field embedding. For
  consistency with previous literature, we preserve ``feature'' in some
  terminologies, e.g., feature interaction, and feature representation.}
vector: \(E=[e_1, e_2, \cdots,e_i, \cdots, e_f]\), where \(f\) denotes
the number of fields, \(e _i \in R^k\) denotes the embedding of \(i\)-th
field , and \(k\) is the dimension of embedding layer.

\subsection{SENET Layer}
As far as we know, different features have various importances for the target task. For
example, the feature occupation is more important than the feature hobby
when we predict a person's income. Inspired by the success of SENET in
the computer vision field, we introduce a SENET mechanism to let the model pay more attention to the feature importance. For specific CTR prediction task, we can
dynamically increase the weights of important features and decrease the
weights of uninformative features via the SENET mechanism.

% rescale xxx embedding by vector a
Using the feature embedding as input, the SENET produces weight vector $A=\{a_1,\cdots,a_i,\cdots,a_f\}$ for field
embeddings and then rescales the original embedding $E$ with vector $A$ to get a new embedding (SENET-Like embedding) $V=[v_1,\cdots, v_i,\cdots,v_f]$, where $a_i \in R$
is a scalar that denotes the weight of the \(i\)-th field
embedding $v_i$, $v_i \in R^k$ denotes the SENET-Like
embedding of \(i\)-th field, \(i \in [1,2,\cdots,f]\) , \(V\) \(\in\)
\(R^{f \times k}\) , \(k\) is an embedding size, and \(f\) is the number
of fields.

\begin{figure}
\centering
\includegraphics[width=8cm, height=5cm]{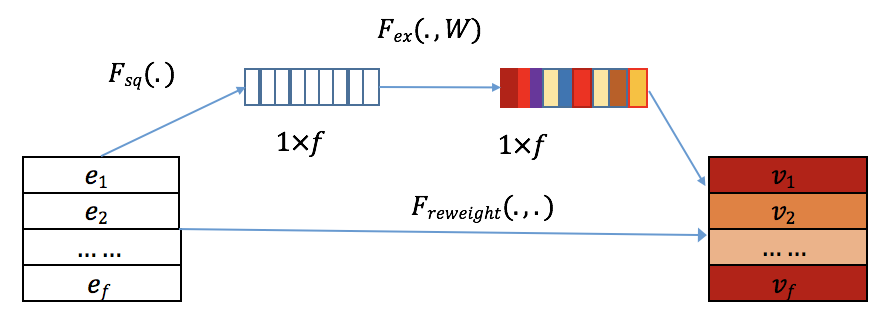}
\caption{The SENET Layer}
\label{fig:f2}
\end{figure}
As Figure \ref{fig:f2} illustrated, the SENET is comprised of three steps: squeeze step, excitation step and re-weight step. These steps can be described in detail as follows:

\noindent\textbf{Squeeze.} This step is used for calculating 'summary statistics'
of each field embedding. Concretely speaking, we use some pooling
methods such as max or mean to squeeze the original embedding
\(E=[e_1,\cdots, e_f]\) into a statistic vector
\(Z  = [z_1,\cdots,z_i, \cdots,  z_f]\), where
\(i \in [1,\cdots,f]\), \(z_i\) is a scalar value which represents the
global information about the \(i\)-th feature representation. \(z_i\)
can be calculated as the following global mean pooling:
\begin{equation}
    z_i = F_{sq}(e_i)=\frac {1}{k}{\sum_{t=1}^{k}e_{i}^{(t)}}
\end{equation}

The squeeze function in original SENET paper\cite{hu2017squeeze} is max pooling. However,
our experimental results show that the mean pooling performs better than
the max pooling. 

\noindent\textbf{Excitation.} This step can be used to learn the weight of each
field embedding based on the statistic vector \(Z\). We use two full
connected (FC) layers to learn the weights. The first FC layer is a
dimensionality-reduction layer with parameters \(W_1\) with reduction
ratio \(r\) which is a hyper-parameter and it uses \(\sigma_1\) as
nonlinear function. The second FC layer increases dimensionality with
parameters \(W_2\). Formally, the weight of field embedding can be
calculated as follows:
\begin{equation}
    A = F_{ex}(Z)=\sigma_2(W_2\sigma_1(W_1Z))
\end{equation}
where \(A \in R^{f}\) is a vector, \( \sigma_1\) and \(\sigma_2\) are
activation functions, the learning parameters are
\( W_1 \in R^{f \times \frac{f}{r}}\),
\(W_2\in R^{\frac{f}{r} \times f}\), and \(r\) is reduction ratio.

\noindent\textbf{Re-Weight.} The last step in SENET is a reweight step which is
called re-scale in original paper\cite{hu2017squeeze}. It does field-wise multiplication
between the original field embedding \(E\) and field weight vector \(A\) and
outputs the new embedding(SENET-Like embedding)
\( V=\{v_1,\cdots,v_i,\cdots,v_f\}\). The SENET-Like embedding \(V\)
can be calculated as follows:
\begin{equation}
V= F_{ReWeight}(A, E)=[a_1\cdot e_1, \cdots, a_f \cdot e_f]=[v_1, \cdots, v_f]
\end{equation}
where \(a_i \in R\), \(e_i \in R^k\), and \(v_i \in R^k\) .

In short, the SENET uses two FCs to dynamically learn the importance of features.
For a specific task, it increases the weights of important
features and decreases the weights of uninformative features.

\subsection{Bilinear-Interaction Layer}

\begin{figure}
\centering
\includegraphics[width=8cm, height=5cm]{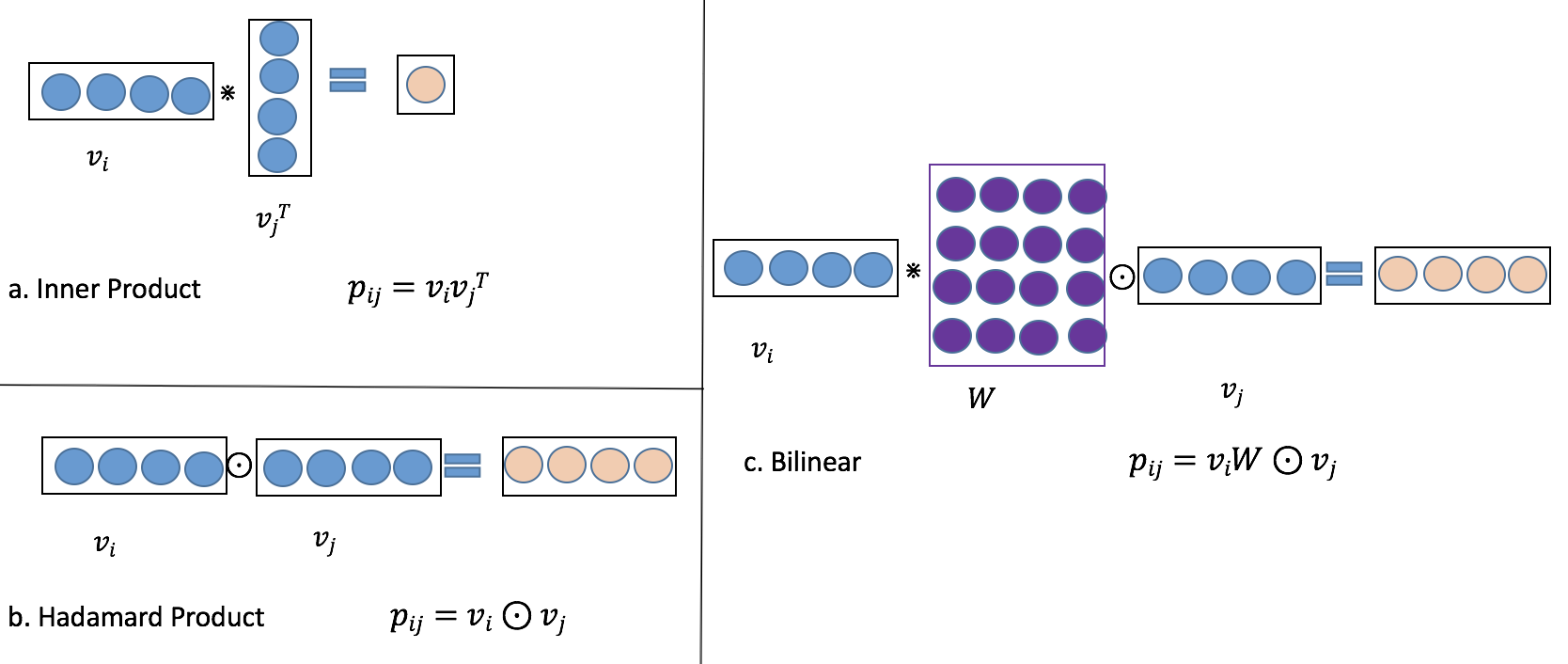}
\caption{The different methods to calculate the feature
interactions. (\textbf{a}): Inner product. (\textbf{b}): Hadamard
product. (\textbf{c}): our proposed bilinear interaction. \textbf{Here
\(p_{ij}\) in inner product method is a scalar while it is a vector in
Hadamard product and our proposed bilinear function.}}
\label{fig:f3}
\end{figure}

The Interaction layer is a layer to calculate the second order feature
interactions. The classical methods of feature interactions in Interaction layer are inner product and Hadamard product. Inner product is widely used in shallow models such as FM and FFM while the Hadamard
product is commonly used in deep models such as AFM and NFM. The forms of inner product and Hadamard product are respectively expressed as \(\{(v_i \cdot  v_j)x_ix_j\}_{(i,j)\in R_x}\)
and \(\{(v_i \odot v_j)x_ix_j\}_{(i,j)\in R_x}\) , where \(R_x=\{(i,j)\}_{i \in \{1,\cdots,f\}, j\in \{1,\cdots,f\}, j >i} \), \(v_i\) is the i-th field embedding vector, \(\cdot\) denotes the regular inner product, and \(\odot\) denotes the Hadamard product, for example, \([a_1, a_2, a_3] \odot [b_1, b_2, b_3] = [a_1b_1, a_2b_2, a_3b_3]\). Inner product and Hadamard product in Interaction layer are too simple to effectively model the feature interactions in sparse dataset. Therefore, we propose a more fine-grained method which combines the inner product and Hadamard product
to learn the feature interactions with extra parameters. As shown in Figure \ref{fig:f3}.c, the inner product is used between the matrix \(W\) and
vector \(v_i\) and the Hadamard product is used between the matrix
\(W\) and the vector \(v_j\). Specifically, we propose three
types of bilinear functions in this layer and call this
layer Bilinear-Interaction layer. Taking the \(i\)-th field embedding
\(v_i\) and the \(j\)-th field embedding \(v_j\) as an example, the
result of feature interaction \(p_{ij}\) can be calculated as follows:

\noindent\textbf{a.Field-All Type}
\begin{equation}
p_{ij}=v_i\cdot W\odot v_j
\end{equation}
where \(W \in R^{k \times k}\), and \(v_i, v_j \in R^k\) are the
\(i\)-th and \(j\)-th field embedding,
\(1 \le i \le f,  i \le j \le f\). Here \(W\) is shared among all
\((v_i, v_j)\) field interaction pairs and there are \(k \times k\)
parameters in Bilinear-Interaction layer, so here we call this type
'Field-All'.

\noindent\textbf{b.Field-Each Type}
\begin{equation}
p_{ij}=v_i \cdot W_i\odot v_j
\end{equation}
where \(W_i \in R^{k \times k}\), \(v_i, v_j \in R^k\) are the \(i\)-th
and \(j\)-th field embedding, \(1 \le i \le f,  i \le j \le f\) . Here
\(W_i\) is the corresponding parameter matrix of the \(i\)-th
field and there are \(f \times k \times k\) parameters in
Bilinear-Interaction layer because we have \(f\) different fields, so
here we call this type 'Field-Each'.

\noindent\textbf{c.Field-Interaction Type}
\begin{equation}
    p_{ij}=v_i \cdot W_{ij}\odot v_j
\end{equation}
where \(W_{ij} \in R^{k \times k}\) is the corresponding parameter matrix
of interaction between field \(i\) and field \(j\) and
\(1 \le i \le f,  i \le j \le f\). The total number of learning
parameters in this layer is \( n \times k \times k\), \(n\) is the
number of field interactions, which is equal to \(\frac{f(f-1)}{2}\).
Here we call this type 'Field-Interaction'.

As shown in Figure \ref{fig:f1}, we have two embeddings(original embedding and SENET-like embedding) and we can adopt either the bilinear function or Hadamard product as feature interaction operation to any embeddings. So we have several combinations of feature interaction in this layer. In Section \ref{sec:s43}, we will discuss the performance of different combinations of bilinear function and Hadamard product in detail. In addition, 
we have three different types of proposed feature interaction methods(Field-All, Field-Each, Field-Interaction) to apply in our model and we will discuss the performance of different field types in Section \ref{sec:s44}. 

In this section, the Bilinear-Interaction layer can output an
interaction vector \(p = [p_1, ..., p_i, ...,p_n]\) from the original
embedding \(E\) and a SENET-Like interaction vector
\(q = [q_1, ..., q_i, ...,q_n]\) from the SENET-Like embedding \(V\), where $p_i, q_i \in R^{k}$ are vectors.

\subsection{Combination Layer}
The combination layer concatenates interaction vector $p$ and $q$ and feeds the concatenated vector into the following layer in
FiBiNET which is a standard neural network layer. It can be expressed
as the following forms:
\begin{equation}
    c = F_{concat}(p,q)=[p_1,\cdots,p_n, q_1,\cdots,q_n]=[c_1,\cdots, c_{2n}]
\end{equation}

If we sum each element in vector \(c\) and then use a sigmoid function
to output a prediction value, we have a shallow CTR model. For further
performance gains, we combine the shallow component and a classical deep
neural network(DNN) which will be described in Section \ref{sec:s35} into a unified model to form the deep network
structure, this unified model is called deep model in our paper.

\subsection{Deep Network}
\label{sec:s35}
The deep network is comprised of several full-connected layers, which implicitly captures
high-order features interactions. As shown in Figure \ref{fig:f1}, the input of deep network
is the output of the combination layer. Let
\(a ^{(0)} = [c_1, c_2, \cdots, c_{2n}]\) denotes the outputs of the
combination layer, where \(c_i \in R^k\) and \(n\) is the number
of field interactions. Then, \(a^{(0)}\) is fed into the deep neural
network and the feed forward process is:
\begin{equation}
    a^{(l)}=\sigma(W^{(l)}a^{(l-1)}+b^{(l)})
\end{equation}
where $l$ is the depth and $\sigma$ is the activation
function. \(W^{(l)}\),\(b^{(l)}\),\(a^{(l)}\) are the model weight, bias
and output of the \(l\)-th layer. After that, a dense real-value feature
vector is generated which is finally fed into the sigmoid function for
CTR prediction: \(y_{d}=\sigma(W^{|L|+1}a^{|L|} + b^{|L|+1})\), where \(|L|\) is the depth of DNN.

\subsection{Output Layer}
To summarize, we give the overall formulation of our proposed model'
output as:
\begin{equation}
    \hat{y}=\sigma(w_0 + \sum_{i=0}^m{w_ix_i}+y_{d})
\end{equation}

where $\hat{y}\in(0,1)$ is the predicted value of CTR, $\sigma$ is
the sigmoid function, $m$ is the feature size, $x$ is a input and
$w_i$ is the $i$-th weight of linear part. The parameters of our model are $\theta=\{w_0, \{w_i\}_{i=1}^{m},\{e_i\}_{i=1}^{m},\{W_i\}_{i=1}^{2},\{W^{(i)}\}_{i=1}^{|L|}\}$.
The learning process aims to minimize the following objective function
(cross entropy):

\begin{equation}
   loss = {-}\frac{1}{N}\sum_{i=1}^{N}({y_i}log(\hat{y_i})+(1-y_i)*log(1-\hat{y_i})) 
\end{equation}
where \(y_i\) is the ground truth of \(i\)-th instance, \(\hat{y_i}\) is
the predicted CTR, and \(N\) is the total size of samples.

\subsubsection{Relationship with FM and FNN} Suppose we remove the SENET layer and Bilinear-Interaction layer, it's not hard to find that our model will be degraded as the FNN. When we further remove the DNN part, and at the same time use a constant sum, then the shallow FiBiNET is downgraded to the traditional FM model.
\section{Experiments}
\label{sec:s4}
In this section, we conduct extensive experiments to answer the
following questions:

\noindent\textbf{(RQ1)} How does our model perform as compared to the
state-of-the-art methods for CTR prediction?

\noindent\textbf{(RQ2)} Can the different combinations of bilinear and Hadamard functions in Bilinear-Interaction layer impact its performance?

\noindent\textbf{(RQ3)} Can the different field types(Field-All, Field-Each and Field-Interaction) of Bilinear-Interaction layer impact its
performance?

\noindent\textbf{(RQ4)} How do the settings of networks influence the performance
of our model?

\noindent\textbf{(RQ5)} Which is the most important component in FiBiNET?

We will answer these questions after presenting some fundamental
experimental settings.
\subsection{Experimental Testbeds and Setup}
\subsubsection{Data Sets} 1) Criteo.
The Criteo\footnote{Criteo
  http://labs.criteo.com/downloads/download-terabyte-click-logs/}
dataset is widely used in many CTR model evaluation. It contains click
logs with 45 millions of data instances. There are 26 anonymous
categorical fields and 13 continuous feature fields in Criteo dataset.
We split the dataset randomly into two parts: 90\% is for training,
while the rest is for testing. 2) Avazu. The Avazu\footnote{Avazu http://www.kaggle.com/c/avazu-ctr-prediction}
dataset consists of several days of ad click-through data which is
ordered chronologically. It contains click logs with 40 millions of data
instances. For each click data, there are 24 fields which indicate
elements of a single ad impression. We split it randomly into two parts:
80\% is for training, while the rest is for testing.

\subsubsection{Evaluation Metrics}
In our experiment, we adopt two metrics: AUC(Area Under ROC) and Log
loss. 

AUC: Area under ROC curve is a widely used metric in evaluating
classification problems. Besides, some work validates AUC as a good
measurement in CTR prediction\cite{graepel2010web}. AUC is insensitive to the
classification threshold and the positive ratio. The upper bound of AUC
is 1, and the larger the better.

Log loss: Log loss is widely used metric in binary classification,
measuring the distance between two distributions. The lower bound of log
loss is 0, indicating the two distributions perfectly match, and a
smaller value indicates better performance.

\subsubsection{Baseline Methods}
To verify the efficiency of combining SENET layer with
Bilinear-Interaction layer in shallow model and deep model, we split our
experiments into two groups: shallow group and deep group. We also split
the baseline models into two parts: shallow baseline models and deep
baseline models. The shallow baseline models include LR(logistic
regression)\cite{mcmahan2013ad}, FM\cite{rendle2010factorization,rendle2012factorization}, FFM\cite{juan2016field,juan2017field}, AFM\cite{xiao2017attentional}, and the
deep baseline models include FNN\cite{zhang2016deep}, DCN\cite{wang2017deep}, DeepFM\cite{guo2017deepfm}, XDeepFM\cite{lian2018xdeepfm}.

Note that an improvement of 1\textperthousand\ in AUC is usually regarded as
significant for the CTR prediction because it will bring a large
increase in a company's revenue if the company has a very large user
base.

\subsubsection{Implementation Details}
\label{sec:414}
We implement all the models with Tensorflow\footnote{TensorFlow:
  https://www.tensorflow.org/} in our experiments. For the embedding
layer, the dimension of embedding layer is set to 10 for Criteo dataset
and 50 for Avazu dataset. For the optimization method, we use the
Adam\cite{kingma2014adam} with a mini-batch size of 1000 for Criteo and 500 for Avazu
datasets, and the learning rate is set to 0.0001. For all deep models, the depth of layers is set to 3, all activation functions are RELU, the number of neurons per layer
is 400 for Criteo dataset and 2000 for Avazu dataset, and the dropout
rate is set to 0.5. For the SENET part, the activation functions in two
FCs are RELU function, and the reduction ratio is set to 3. We conduct
our experiments with 2 Tesla K40 GPUs.

\subsection{Performance Comparison(RQ1)}
In this subsection, we summarize the overall performance of shallow
models and deep models on Criteo and Avazu test sets in Table \ref{table:t1} and Table \ref{table:t2} respectively.

\begin{table}[!htp]
  \caption{The overall performance of shallow models on Criteo and
Avazu datasets. The SE-FM-ALL denotes the shallow model with the
Field-All type of Bilinear-Interaction layer.}
  \label{table:t1}
  \begin{tabular}{lcccr}
    \toprule
     &\multicolumn{2}{c}{Criteo}&\multicolumn{2}{c}{Avazu}\\
    \hline
     Model & AUC & Logloss & AUC & Logloss\\
    \midrule
LR & 0.7808 & 0.4681 & 0.7633 & 0.3891\tabularnewline
FM & 0.7923 & 0.4584 & 0.7745 & 0.3832\tabularnewline
FFM & 0.8001 & 0.4525 & 0.7795 & 0.3810\tabularnewline
AFM & 0.7965 & 0.4541 & 0.7740 & 0.3839\tabularnewline
\textbf{SE-FM-All} & \textbf{0.8021} & \textbf{0.4495} &
\textbf{0.7803} & \textbf{0.3800}\tabularnewline
  \bottomrule
\end{tabular}
\end{table}

% Table \ref{table:t1} on page
Table \ref{table:t1} shows the results of the shallow models on Criteo and Avazu
datasets. We find our shallow SE-FM-All model consistently
outperforms other models such as FM, FFM, AFM etc. On the one hand, the
results indicate that combining the SENET
mechanism with the bilinear interaction over sparse features is an
effective method for many real world datasets; on the other hand, for the classical shallow models,
the state-of-the-art model is FFM which is restricted by the need of large memory
and cannot be easily used in Internet companies, our shallow model has
fewer parameters but still performs better than FFM. So it can be
regarded as an alternative solution for FFM.

\begin{table}[!htp]
  \caption{The overall performance of deep models on Criteo and
Avazu datasets. The DeepSE-FM-ALL denotes the deep model with the
Field-All type of Bilinear-Interaction layer.}
  \label{table:t2}
  \begin{tabular}{lclcr}
    \toprule
     &\multicolumn{2}{c}{Criteo}&\multicolumn{2}{c}{Avazu}\\
    \hline
     Model& AUC & Logloss & AUC & Logloss\\
    \midrule
FNN & 0.8057 & 0.4464 & 0.7802 & 0.3800\tabularnewline
DeepFM & 0.8085 & 0.4445 & 0.7786 & 0.3810\tabularnewline
DCN & 0.7978 & 0.4617 & 0.7681 & 0.3940\tabularnewline
XDeepFM & 0.8091 & 0.4461 & 0.7808 & 0.3818\tabularnewline
\textbf{DeepSE-FM-All} & \textbf{0.8103} & \textbf{0.4423} &
\textbf{0.7832} & \textbf{0.3786}\tabularnewline
  \bottomrule
\end{tabular}
\end{table}

For further performance gains, we combine the shallow part and DNN into
a deep model. The overall performance of deep models is shown in Table \ref{table:t2} and we have the following observations:

\begin{itemize}
\item
  Combining the shallow part and DNN into a unified model, the shallow
  model can gain further performance improvement. We can infer from experimental results that the implicit high-order feature interactions help the shallow model to gain
  more expressive power.
\item
  Among all the compared methods, our proposed deep FiBiNET achieves the best performance. Our deep model outperforms FNN by relatively 0.571\% and 0.386\% in terms of AUC(0.918\% and 0.4\% in terms of log loss) and outperforms DeepFM by 0.222\%
  and 0.59\% in terms of AUC(0.494\% and 0.6\% in terms of log loss) on Criteo and Avazu datasets.
\item
  The results indicate that combining the SENET mechanism with Bilinear-Interaction in DNN for prediction is effective. On the one hand, the SENET intrinsically introduces dynamics conditioned on the input, helping to boost feature discriminability; on the other hand, the
  bilinear function is an effective method to model the feature interaction as compared with other methods such as the inner product or Hadamard product as described in Section \ref{sec:s43}.
\end{itemize}

For further performance gains, we will discuss the different combinations
of Bilinear-Interaction layer in Section \ref{sec:s43} and the field types of
Bilinear-Interaction layer in Section \ref{sec:s44}.

\subsection{Combinations of Bilinear-Interaction Layer(RQ2)}
\label{sec:s43}
In this section, we will discuss the influence of different type of combinations between bilinear function and Hadamard product in Bilinear-Interaction layer. For convenience, we use 0 and 1 to represent which function is used in Bilinear-Interaction layer. The '1' denotes that bilinear function is used while 0 means Hadamard product is used. We have two embeddings so two numbers are used. The first number denotes the feature interaction method used on original embedding and the second number denotes the feature interaction method used on SENET-Like embedding. For example, '10' denotes that bilinear function is used as feature interaction method on the original embedding while the Hadamard function is used as feature interaction method on the SENET like embedding. Similarly, we conduct the experiments on shallow and deep models and summarize the results in Table \ref{table:t3}.

\begin{table}[!htp]
  \caption{The performance of different combinations of bilinear and Hadamard functions in Bilinear-Interaction layer. The field type of Bilinear-Interaction layer is set to Field-Each.}
  \label{table:t3}
  \begin{tabular}{lclcr}
    \toprule
     &\multicolumn{2}{c}{Criteo}&\multicolumn{2}{c}{Avazu}\\
    \hline
     Combinations & AUC & Logloss & AUC & Logloss\\
    \midrule
SE-FM\_00 & 0.7989 & 0.4525 & 0.7782 & 0.3818\tabularnewline
SE-FM\_01 & 0.8018 & 0.4500 & \textbf{0.7797} & 0.3808\tabularnewline
SE-FM\_10 & 0.8029 & 0.4488 & 0.7794 & \textbf{0.3807}\tabularnewline
SE-FM\_11 & \textbf{0.8037} & \textbf{0.4479} & 0.7770 & 0.3815\tabularnewline
\hline
DeepSE-FM-00 & \textbf{0.8105} & \textbf{0.4425} & 0.7828 & 0.3785\tabularnewline
DeepSE-FM-01 & 0.8104 & 0.4423 & \textbf{0.7833} &
\textbf{0.3783}\tabularnewline
DeepSE-FM-10 & 0.8100 & 0.4427 & 0.7810 & 0.3809\tabularnewline
DeepSE-FM-11 & 0.8099 & 0.4428 & 0.7805 & 0.3807\tabularnewline
  \bottomrule
\end{tabular}
\end{table}
Overall, we can not draw any obvious conclusions, but we can find some
empirical observations as follows:

\begin{itemize}
\item
  On Criteo dataset, the combination '11' outperforms other type of combinations among
  shallow models. However, the combination '11' performs worst among the
  deep models.
\item
  The preferred combination in deep models should be '01'. This combination means the
  bilinear function is only applied to the SENET-Like embedding layer,
  which is beneficial for designing an effective network architectures
  in our model.
\end{itemize}

\subsection{Field Types of Bilinear-Interaction (RQ3)}
\label{sec:s44}
\begin{table}[!htp]
  \caption{The performance of different field types of
Bilinear-Interaction layer.}
  \label{table:t4}
  \begin{tabular}{lclcr}
    \toprule
     &\multicolumn{2}{c}{Criteo}&\multicolumn{2}{c}{Avazu}\\
    \hline
     Field Types & AUC & Logloss & AUC & Logloss\\
    \midrule
SE-FM-All & 0.8021 & 0.4495 & \textbf{0.7804} & \textbf{0.3800}\tabularnewline
SE-FM-Each & 0.8037 & 0.4479 & 0.7797 & 0.3812\tabularnewline
SE-FM-Interaction & \textbf{0.8059} & \textbf{0.4460} & 0.7785 &
0.3815\tabularnewline
\hline
DeepSE-FM-All & 0.8103 & 0.4423 & 0.7832 & 0.3786\tabularnewline
DeepSE-FM-Each & 0.8104 & 0.4423 & \textbf{0.7833} &
\textbf{0.3783}\tabularnewline
DeepSE-FM-Interaction & \textbf{0.8105} & \textbf{0.4421} & 0.7828 &
0.3788\tabularnewline
  \bottomrule
\end{tabular}
\end{table}

In this section, we study impact of different field types(Field-All, Field-Each and Field-Interaction) of Bilinear-Interaction layer. We
first fix the combination of the bilinear and Hadamard product in
Bilinear-Interaction layer. The combination of Bilinear-Interaction
layer is set to '01' for deep model and '11' for shallow model. And the
mark '01' and '11' are illustrated in Section \ref{sec:s43}. We summarize the experimental results in
Table \ref{table:t4} and have the following observations:
\begin{itemize}
\item
  For the shallow models, compared to the Field-All type of our shallow model (in Table \ref{table:t1}),
  the Field-Interaction type can gains
  0.382\% (relatively 0.476\%) improvements in terms of AUC on Criteo dataset.
\item
  For the deep models, compared to the Field-All type of our deep model (in Table
  \ref{table:t2}), the type of Field-Interaction for Criteo dataset and Field-Each for Avazu dataset can gain some improvements respectively.
\item
  The performances of different types of Bilinear-Interaction layer depend on datasets. On Criteo dataset, the performance ranking is as follows: Field-Interaction, Field-Each, and Field-All. While on Avazu dataset, we cannot draw the obvious conclusion. 
\end{itemize}

\subsection{Hyper-parameter Investigation(RQ4)}
In this subsection, we will conduct some hyper-parameter investigations in our model. We focus on hyper-parameters in the following two components in FiBiNET: the embedding part and the DNN part. Specifically, we change the following hyper-parameters:(1) the dimension of
embeddings; (3) the number of
neurons per layer in DNN; (4) the depth of DNN. Unless specially
mentioned in our paper, the default parameter of our network is set as
the Section \ref{sec:414}.

\subsubsection{Embedding Part}
\begin{table}[!htp]
  \caption{The performance of different embedding sizes on
Criteo and Avazu datasets}
  \label{table:t5}
  \begin{tabular}{lclcr}
    \toprule
     &\multicolumn{2}{c}{Criteo}&\multicolumn{2}{c}{Avazu}\\
    \hline
     Embedding-Size & AUC & Logloss & AUC & Logloss\\
    \midrule
10 & \textbf{0.8104} & \textbf{0.4423} & 0.7809 & 0.3801\tabularnewline
20 & 0.8093 & 0.4435 & 0.7810 & 0.3796\tabularnewline
30 & 0.8071 & 0.4460 & 0.7812 & 0.3799\tabularnewline
40 & 0.8071 & 0.4464 & 0.7824 & 0.3790\tabularnewline
50 & 0.8072 & 0.4468 & \textbf{0.7833} & \textbf{0.3787}\tabularnewline
  \bottomrule
\end{tabular}
\end{table}
We change the embedding sizes from 10 to 50 and summarize the experimental results in Table \ref{table:t5}. We can find some observations as follows:

\begin{itemize}
\item
  As the dimension is expanded from 10 to 50, our model can obtain a substantial improvement on Avazu dataset. 
\item
  The performance degrades when we increase the embedding size on Criteo dataset. Enlarging embedding size indicates increasing the number of parameters in embedding layer and DNN part.
  We guess that it may be the much more features in Criteo dataset as opposed to Avazu dataset that leads to optimization difficulties.
\end{itemize}

\subsubsection{DNN Part}
In deep part, we can change the number of neurons per layer, the depths
of DNN, the activation functions and the dropout rates. For brevity, we
just study the impact of different neural units per layer and different depths in DNN part.

% \begin{figure}
% \centering
% \includegraphics[width=8cm, height=5cm]{image-20181025003821109.png}
% % \includegraphics[width=9cm, height=5cm]{image-20181025160427429.png}
% % \includegraphics{image-20181020000624735.png}
% \caption{The performance of different number of layers in DNN.}
% \label{fig:f4}
% \end{figure}
\begin{figure}
\includegraphics[width = 4cm]{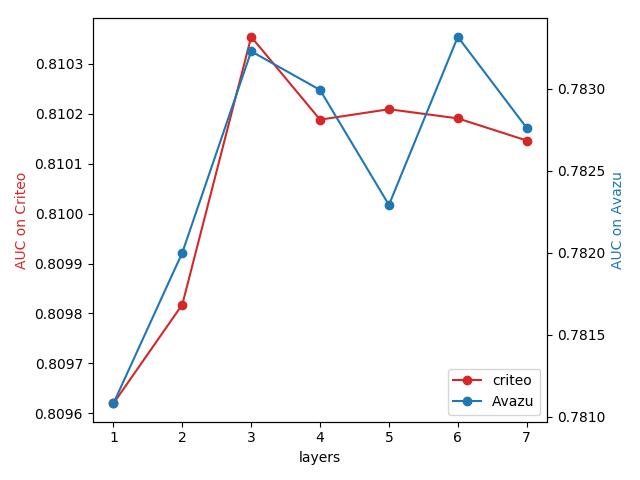}  
\includegraphics[width = 4cm]{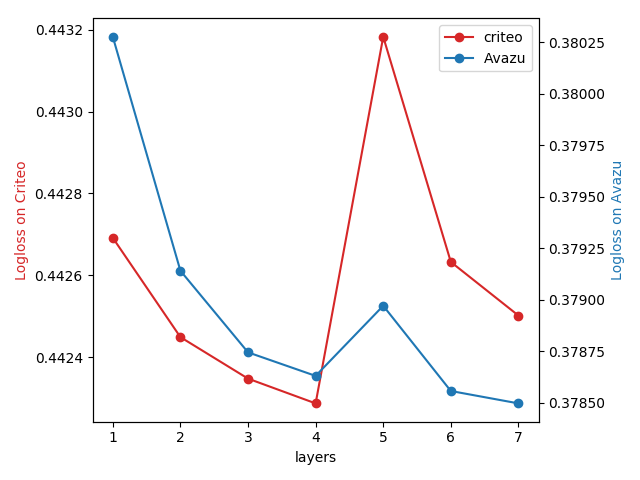}
% \includegraphics[width = 4cm]{dnn_neurons.jpg}  
% \includegraphics[width = 4cm]{dnn_neurons_loss.jpg}
% \subfloat[fig 3]{\includegraphics[width = 3cm]{dnn_neurons.jpg}}
% \subfloat[fig 4]{\includegraphics[width = 3cm]{dnn_neurons_loss.jpg}} 
\caption{The performance of different number of layers in DNN.}
\label{fig:f4}
\end{figure}
\begin{figure}
\includegraphics[width = 4cm]{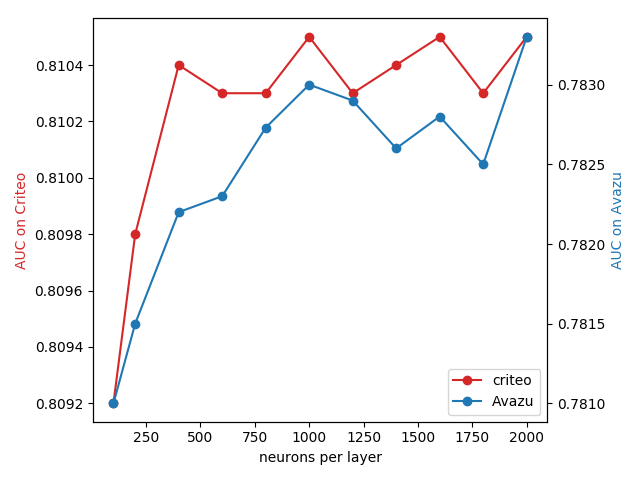}  
\includegraphics[width = 4cm]{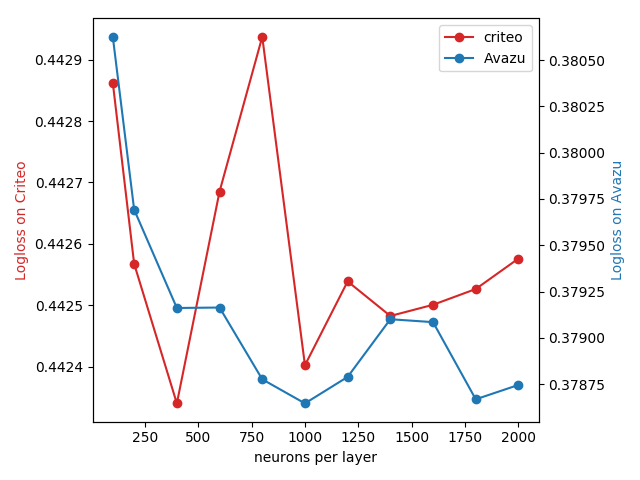}
% \subfloat[fig 3]{\includegraphics[width = 3cm]{dnn_neurons.jpg}}
% \subfloat[fig 4]{\includegraphics[width = 3cm]{dnn_neurons_loss.jpg}} 
\caption{The performance of different number of neurons per layer in DNN}
\label{fig:f5}
\end{figure}

As a matter of fact, increasing the number of layers can increase the
model complexity. We can observe from Figure \ref{fig:f4} that increasing number of layers improves model performance at the beginning. However, the
performance is degraded if the number of layers keeps increasing. This
is because an over-complicated model is easy to overfit. It's a good
choice that the number of hidden layers is set to 3 for Avazu dataset
and Criteo dataset. Likewise, increasing the number of neurons per layer introduces complexity. In Figure \ref{fig:f5}, we find that it is better to set 400 neurons per layer for Criteo dataset and 2000 neurons per layer for Avazu dataset.

% \begin{figure}
% \centering
% \includegraphics[width=8cm, height=5cm]{image-20181025110512959.png}
% % \includegraphics[width=9cm, height=5cm]{image-20181025160427429.png}
% % \includegraphics{image-20181020000624735.png}
% \caption{The performance of different number of neurons per layer in DNN.}
% \label{fig:f5}
% \end{figure}

\subsection{Ablation Study (RQ5)}

\label{sec:s5}
\begin{table}[!htp]
\centering
  \caption{The performance of different components in FiBiNET.}
  \label{table:t51}
  \begin{tabular}{lclcr}
    \toprule
     &\multicolumn{2}{c}{Criteo}&\multicolumn{2}{c}{Avazu}\\
    \hline
     Model & AUC & Logloss & AUC & Logloss\\
    \midrule
BASE & 0.8037 & 0.4479 & 0.7797 & 0.3812\tabularnewline
NO-SE & 0.7962 & 0.4552 & 0.7763 & 0.3825\tabularnewline
NO-BI & 0.7986 & 0.4525 & 0.7754 & 0.3829\tabularnewline
FM & 0.7923 & 0.4584 & 0.7745 & 0.3832\tabularnewline
\hline
Deep-BASE & 0.8104 & 0.4423 & 0.7833 &
0.3783\tabularnewline
NO-SE & 0.8098 & 0.4427 & 0.7822 & 0.3790\tabularnewline
NO-BI & 0.8093 & 0.4435 & 0.7827 & 0.3785\tabularnewline
% DeepFM & 0.8065 & 0.4445 & 0.7786 & 0.3810\tabularnewline
FNN & 0.8057 & 0.4464 & 0.7802 & 0.3800\tabularnewline
  \bottomrule
\end{tabular}
\end{table}
Although we have demonstrated strong empirical results, the results presented so far have not isolated the specific contributions from each component of the FiBiNET.\ In this section, we perform ablation experiments over FiBiNET in order to better understand their relative importance. We set `DeepSE-FM-Interaction' as the base model and perform it in the following ways:
1) \textbf{No BI}:\ remove the Bilinear-Interaction layer from FiBiNET 
2)  \textbf{No SE}:\ remove the SENET layer from FiBiNET.

If we remove the SENET layer and Bilinear-Interaction layer, our shallow FiBiNET and deep FiBiNET will downgrade to FM and FNN. We can find the following observations in Table \ref{table:t51}:
\begin{itemize}
\item Both the Bilinear-Interaction layer and SENET layer are necessary for FiBiNET's performance. We can see that the the performance will drop apparently when we remove any component. 
\item The Bilinear-Interaction layer is as important as the SENET layer in FiBiNET. 

\end{itemize}

\section{Conclusions}
\label{sec:s5}
Motivated by the drawbacks of the state-of-the-art models, 
we propose a new model named FiBiNET as an abbreviation for Feature Importance and Bilinear feature Interaction NETwork and aim to dynamically learn the feature importance and fine-grained feature interactions. Our proposed FiBiNET makes a contribution to
improving performance in these following aspects: 1) For CTR task, the
SENET module can learn the importance of features dynamically. It boosts
the weight of the important feature and suppresses the weight of
unimportant features. 2) We introduce three types of
Bilinear-Interaction layers to learn feature interaction rather than
calculating the feature interactions with Hadamard product or inner
product. 3) Combining the SENET mechanism with bilinear feature
interaction in our shallow model outperforms other shallow models such
as FM and FFM. 4) In order to improve performance further, we combine a classical deep neural network(DNN) component with the shallow model to be a deep model. The deep FiBiNET consistently outperforms the other state-of-the-art deep models such as DeepFM and XdeepFM.

\bibliographystyle{ACM-Reference-Format}
\bibliography{sample-base}
\end{document}